\documentclass[final]{anthology-ch}

\usepackage{booktabs}
\usepackage{multirow}

\usepackage{graphicx}
\usepackage{float}
\usepackage{todonotes}

\usepackage[T1]{fontenc}
\usepackage{lmodern}

\usepackage{caption}
\usepackage{subcaption}  
\usepackage{enumitem}
\usepackage{makecell}
\usepackage{graphicx}
\usepackage{linguex}

\title{Modeling the Construction of a Literary Archetype: The Case of the Detective Figure in French Literature}

\author[1]{Jean Barré}[orcid=0000-0002-1579-0610]
\author[1]{Olga Seminck}[orcid=0000-0003-4617-5992]
\author[1]{Antoine Bourgois}[orcid=0009-0008-8117-4120]
\author[1]{Thierry Poibeau}[orcid=0000-0003-3669-4051]

\affiliation{1}{LaTTiCe Laboratory, CNRS-ENS-PSL, Paris, France}


\pubyear{2025}
\pubvolume{1}
\pagestart{1}
\pageend{1}
\conferencename{Proceedings of Computational Humanities Research 2025}
\conferenceeditors{Editor1 Editor2}
\doi{00000/00000}

\begin{document}

\maketitle


\begin{abstract}
This research explores the evolution of the detective archetype in French detective fiction through computational analysis. Using quantitative methods and character-level embeddings, we show that a supervised model is able to capture the unity of the detective archetype across 150 years of literature, from \textit{M. Lecoq} (1866) to \textit{Commissaire Adamsberg} (2017). Building on this finding, the study demonstrates how the detective figure evolves from a secondary narrative role to become the central character and the “reasoning machine” \parencite{symons_bloody_1993} of the classical detective story. In the aftermath of the Second World War, with the importation of the hardboiled tradition into France, the archetype becomes more complex, navigating the genre's turn toward social violence and moral ambiguity.
\end{abstract}


\section{Introduction}

Distant reading \cite{moretti2013} has become a key methodological approach in literary studies, enabling the large-scale analysis of digitized corpora through computational techniques. By moving beyond the traditional canon to include lesser-studied works, it offers a broader perspective on literary trends over time \cite{moretti2000conjectures, underwood2019distant, barre_operationalizing_2023, barre_latent_2024}. While distant reading can be applied to various aspects of literary texts, the analysis of literary characters has emerged as a particularly productive area of research, allowing scholars to explore narrative structures, character portrayals, and cultural representations at scale.

A foundational work in this line of inquiry is provided by \textcite{bamman_bayesian_2014}, who propose a Bayesian mixed-effects model to infer latent character types from a large corpus of English novels. Their study demonstrates that computational methods can identify underlying character personas by analyzing linguistic patterns associated with character descriptions, actions, and dialogues. Following this seminal approach, the computational literary studies community has developed specialized tools to automate the extraction of character information. Notable examples include BookNLP for English \cite{bamman-etal-2020-annotated,github_bamman}, as well as its counterparts adapted to other languages such as French \cite{booknlp_fr_2024}, German \cite{ehrmanntraut-etal-2023-llpro}, and Dutch \cite{van2023putting}. These pipelines enable researchers to systematically identify and cluster all mentions of a given character, alongside associated linguistic features including agentive and passive verbs, modifiers, and direct speech. Leveraging these computational methods within distant reading frameworks, subsequent research has explored significant literary phenomena, notably the representation of gender \cite{underwood_transformation_2018,naguib2022romanciers,vianne_gender_2023} and the structural analysis of plot dynamics \cite{konle_modeling_2022,konle_character_2023}. In this study, we use this character modeling approaches to study the emergence and evolution of the detective archetype in French fiction.

The figure of the detective has become a ubiquitous presence in contemporary popular culture, extending far beyond literature into cinema, television series, and various media adaptations. However, its roots lie firmly within English and French literary traditions \cite[12]{brownson_figure_2014}, particularly between the late 19th and early 20th centuries, a period crucial to the construction and popularization of detective fiction as a genre. 

We decided to focus exclusively on the detective archetype in French literature. Although there are numerous interconnections and reciprocal influences with the Anglo-American tradition, illustrated by Poe's Dupin inspiring Gaboriau's Lecoq \cites[43]{schutt_rivalry_1998}{Pandzic_2020}, who in turn influenced Conan Doyle's Sherlock Holmes \cites[394]{bonniot_emile_1985}[110]{lycett_man_2007}, the depth and distinctiveness of the French detective tradition alone present a substantial challenge.



To be more specific, we build a classifier that is able to automatically detect characters that are detectives (Section~\ref{section:detective_detector}). We study the emergence of the detective figure by quantifying the percentage of detectives per year in a large corpus of French fiction \textit{Chapitres} Corpus \cite{leblond_2022_7446728}, and show that the detective characters tends to be more and more central to the novels as the literary genre is building up (Section~\ref{section:emergence_centrality}). 
Finally, we also demonstrate that our methods support literary theory about the emergence of detective subgenres, such as the \textit{Hard-Boiled} detectives (putting forward solitary characters fighting organized crime as well as corrupt authorities) and French \textit{Roman Noir} or \textit{Neo-Polar} (Section~\ref{section:semantic_trajectory}). Our computational analysis shows how the critical perspective from canonical whodunit figures expands to broader representations of the detective, capturing subtle shifts and variations in characterization over time. 


\section{The Detective Archetype}

\subsection{Definition}

The detective archetype constitutes a foundational figure within detective fiction, evolving from a peripheral puzzle solver to a central character embodying rationality and methodical reasoning. As Dubois emphasizes, classic detectives such as Dupin, Holmes, and Rouletabille favor purely intellectual work, distancing themselves from direct action: \enquote{he solves the problems presented to him by pure analytical deduction} \parencite{dubois_naissance_1985}. This approach aligns closely with the notion of a \textit{reasoning machine} \parencite{symons_bloody_1993}, which disregards human motives or psychology, focusing solely on the accuracy of deductions regarding actions. Consequently, the detective evolves into an extreme symbol of rationality, whose methods rely systematically on interpreting clues and signs to solve a mystery.

This fundamental definition of the detective is intimately linked to the \textit{whodunit}, a subgenre of detective fiction that stands out for its puzzle-like plot, with clues and red herrings carefully scattered to invite the reader to match wits with the detective and try to uncover the truth before the final revelation. Todorov \cite{todorov_typologie_1980} analyzes the specificity of the \textit{whodunit} by highlighting a double narrative structure: the hidden story of the crime and the visible story of the investigation. In this way, the identity and role of the detective are inseparable from the \textit{whodunit}: the detective becomes both the reader's guide and double in logically reconstructing a crime that has already occurred.

While canonical detective figures such as Doyle's \textit{Sherlock Holmes} or Gaboriau's \textit{Père Tabaret} epitomize this rational genius archetype, meaning that intelligence and ingenious reasoning are key to solving the mystery by the end of the novel, the archetype is not limited to officially sanctioned investigators. \textit{Holmes} or \textit{Tabaret} themselves, for instance, are no professional police detectives. Rather, the archetype comprises a complex aggregation of characters, roles, and traits that evolve through time and across literary subgenres. It can be embodied by a wide variety of characters, ranging from private and amateur detectives to police officers, overly inquisitive journalists, and even criminals \cite{ALL_detective_2012}.

In France, this diversification materialized through increasingly ambiguous and complex figures, such as Arsène Lupin, simultaneously detective and criminal, pursued by genuine detectives like Commissioner Ganimard or the amateur sleuth Isidore Beautrelet. These diverse incarnations underscore the archetype's adaptability and illustrate how detective fiction explores multiple perspectives, social roles, and investigative approaches through a spectrum of distinct yet interconnected characters.

\subsection{Origins}

The detective archetype emerged from mid-19th-century literary developments, influenced by proto-detective figures such as Edmond Dantès (\textit{Le Comte de Monte Cristo}), Rodolphe (\textit{Les Mystères de Paris}), and Rocambole (Ponson du Terrail) \parencite{lavergne_naissance_2009}. Its origins were also shaped by figures like Eugène François Vidocq \parencite{gerson_vidocq_1977}, a former criminal who became the first chief of the \textit{Sûreté} in 1811 and founded the first modern detective agency, \textit{Le Bureau de Renseignements}. Vidocq directly inspired Balzac's Vautrin, and Poe's Dupin, two proto‑detective figures.


The foundation of the detective novel was laid by Edgar Allan Poe \cite{ALL_detective_2012}, through three short stories featuring Auguste Dupin, a non-professional detective who solves crimes through a combination of intuitive reasoning and scientific tools: a method Poe called \textit{ratiocination} \cite{murray_thesis_on_Poe}. While Edgar Allan Poe originally formulated the detective formula in short story form \cite[81]{cawelti_adventure_1997}, Émile Gaboriau (1832-1873) was the first to adapt it to the novel format within French tradition.
  
The genre continued to evolve with figures such as \textit{Rouletabille} by Gaston Leroux, Maurice Leblanc's \textit{Arsène Lupin}, and later Georges Simenon's \textit{Maigret}. This diversity of figures illustrates a progressive construction of the archetype through successive layers, shaped by varied sociocultural and literary contexts.

Detective fiction gradually established itself as a major literary genre by the late 19th century and, by the 1920s–30s, had become a central reference in the collective imagination. \textcite{messac_detective_2011}, among the first to study the genre systematically, defined it as \enquote{a narrative dedicated above all to the methodical and gradual discovery, through rational means, of the exact circumstances of a mysterious crime}. Its specificity lies in a narrative structure that moves from an initial crime or mystery to its reconstruction, generating dramatic tension through multiple suspects and building suspenseful reader expectations \cite{ALL_detective_2012}. Beyond suspense, the detective is pivotal: their reasoning drives the story toward resolution \cite{williard_huntington_wright_great_1947}. It is through the detective's logic, insight, and analytical skill that the plot unfolds.


\section{The Detective Detector}\label{section:detective_detector}

Building on this rich panorama, our goal is to extract the detective archetype as a common matrix across all these figures, from \textit{Père Tabaret} (first appearence in 1866) to \textit{Commissaire Adamsberg} (last appearence in 2017), and, paradoxically, as an entity that fully coincides with none of them. We aim to implement a quantitative approach\footnote{Data and code of this paper available at \url{https://github.com/lattice-8094/DETECTIVES}} capable of identifying, in each text, the behavioral, narrative, and linguistic invariants that underlie the investigator's stance, despite the diversity of eras, styles, and roles (methodical, intuitive, empathetic, cynical, ...).


\subsection{Corpus and Data Annotation}

This study is based on a large corpus of French fiction, the \textit{Chapitres} Corpus \cite{leblond_2022_7446728}, which includes nearly 3,000 texts, including a lot of detective fiction novels. 

Detecting the detective archetype during 150 years of the detective genre is by no means trivial. It requires two things: first, the ability to identify characters reliably at the scale of a novel \cite{Bourgois2025, Barré2023}. That is done using the Propp-fr pipeline\footnote{\url{https://github.com/lattice-8094/propp} (accessed July 11, 2025).}, an extension of BookNLP-fr \cite{booknlp_fr_2024} that facilitates automatic coreference chain recognition at the scale of entire novels (over 100k tokens). Second, the ability to distinguish, among these characters, which are detectives and which are not. 

Our annotation aimed to represent the detective archetype across various subgenres, from the French \textit{whodunit} to \textit{série noire} thrillers. We focused on paradigmatic figures with significant influence due to their recurring roles: amateur journalists (\textit{Rouletabille}, Gaston Leroux, 1907), outlaw vigilantes (\textit{Arsène Lupin}, Maurice Leblanc, 1908), detective-crooks (\textit{Larsan}, from the same novel as \textit{Rouletabille}), honest professionals (\textit{Lecoq}, Emile Gaboriau; \textit{Maigret}, Georges Simenon), and French adaptations of hard-boiled detectives (\textit{Nestor Burma}, Léo Malet, 1942; \textit{Gabriel Lecouvreur}, Didier Daeninckx, 1981; \textit{Adamsberg}, Fred Vargas, 1991).

In total, 185 characters across 156 novels were annotated as \textit{Detectives}, representing 47 unique figures (some appear in multiple works). We also annotated 419 characters as \textit{Non-Detectives}. Since our goal is binary classification (\textit{detective} vs.\ \textit{not detective}), we ensured a diverse negative set so the model could learn what does not define the archetype. To this end, we randomly selected 419 non-detective characters from the Chapitres corpus, stratified by time to reflect a variety of roles and narrative functions.

We also included characters from the mid-19th century, prior to the emergence of the genre, to expose the model to periods where the \textit{detective figure} as we know it had yet to appear.

\subsection{Character Embeddings}

Once the characters are annotated as Detective or Non-Detective, we created vectorized representations of them that we will refer to as \textit{character embeddings}. These semantic representations are built upon the vocabulary extracted alongside the coreference chains by the BookNLP-fr tool. For each character, we retrieve the actions they perform or endure (grammatically agent and passive verbs), modifiers (such as adjectives and other predicatives), and nouns that come after a possessive determiner. We hypothesize that the detective archetype involves a specific textual characterization that can thus be captured quantitatively.

These elements are used as the basis for the semantic representation. Trying to represent each character in a vector space, we implemented two distinct approaches.

\paragraph{Bag of Words representation:}
 The first is a baseline with Bag-of-Words (BoW). The idea is to control the dimensions of the vector space by simply selecting the 1,000 Most Frequent Words (MFW) from the character attributes retrieved by BookNLP-fr. For each character in our annotated dataset, we retrieved the relative frequency, of every MFW associated with the character.

\paragraph{Contextual Embeddings with CamemBERT:}

The second method is based on contextualized word embeddings generated by the CamemBERT\textsubscript{LARGE} encoder-only language model \cite{martin2019camembert}. For each attribute linked to a character, we gather contextual embeddings.
The final character embedding vector is obtained by averaging the contextual embeddings of all attributes linked to that character.
This element-wise averaging yields a dense semantic representation (1,024 dimensions) that captures the character's descriptive context beyond surface lexical frequency.
Passive verbs (those for whom the character is the object) were excluded from this process, as preliminary experiments showed they provided limited discriminative power for the classification task due to their sparseness.


\subsection{Supervised Classification}

To predict whether a character is a detective, we used manually annotated gold-standard labels as ground truth and fed precomputed character embeddings into models. We compared two Scikit-learn classifiers \cite{scikit-learn}: Support Vector Machines (SVM) and Logistic Regression.

To handle class imbalance, we used stratified 5-fold cross-validation (\texttt{StratifiedKFold}), maintaining the detective/non-detective ratio across splits. Model performance was evaluated using balanced accuracy (\texttt{balanced\_accuracy\_score}).

To prevent information leakage (e.g., from characters like inspector Maigret appearing in both training and test sets) we applied a \texttt{Leave-One-Group-Out} (LOGO) strategy. When grouping by author, all books by the same author were placed entirely in either the training or test set. We define three grouping schemes:
\begin{enumerate}[itemsep=0.25em, topsep=0pt]
\item \emph{Character:} all instances of the same character are held out together.
\item \emph{Author:} all characters by the same author are placed in one group.
\item \emph{Time Period:} all characters originating from the same historical era form a group.
\end{enumerate}



\subsection{Quantitative Evaluation}

Table \ref{tab:model_results_detailed} reports balanced accuracy alongside precision, recall, and F1-scores for both classes under our three main feature–model configurations, as well as under the strict LOGO validation schemes. Starting from the simple Bag-of-Words baseline, Logistic Regression achieves a balanced accuracy of 0.836, but struggles particularly on the detective class (F1 = 0.75). Replacing the classifier with an SVM yields a substantial jump (B. Acc. = 0.895), with both precision and recall improving across classes; a testament to the SVM's capacity to carve sharper decision boundaries even on sparse, high-dimensional vectors.

Moving to contextual embeddings, CamemBERT paired with Logistic Regression already outperforms both BoW setups (B. Acc.=0.906), delivering a notable boost in recall for detectives (0.919). Finally, when combining CamemBERT embeddings with an SVM, we observe the strongest results overall: a balanced accuracy of 0.923, precision and F1-scores near or above 0.94 for non-detectives, and an F1 of 0.887 for detectives. This configuration does not only maximize our core metric but also maintains high precision and recall on the under-represented detective class, minimizing both false positives and false negatives.

Given these results, we will adopt the CamemBERT + SVM model for all subsequent experiments.


\begin{table}[H]
\centering
\begin{tabular}{l c c c c c c c}
\toprule
\multirow{2}{*}{\textbf{Model}} 
  & \multirow{2}{*}{\textbf{B. Acc.}} 
  & \multicolumn{3}{c}{\textbf{Non-detective}} 
  & \multicolumn{3}{c}{\textbf{Detective}} \\
\cmidrule(lr){3-5}\cmidrule(lr){6-8}
 &  & \textbf{P} & \textbf{R} & \textbf{F1-score} 
    & \textbf{P} & \textbf{R} & \textbf{F1-score} \\
\midrule
BoW + LogReg       & 0.836 & 0.94  & 0.78  & 0.85  & 0.64 & 0.89 & 0.75  \\ 
BoW + SVM          & 0.895 & 0.95  & 0.91  & 0.93  & 0.81 & 0.88 & 0.84  \\ 
CamemBERT + LogReg & 0.906 & \textbf{0.961} & 0.893 & 0.926 & 0.791 & \textbf{0.919} & 0.850   \\ 
\textbf{CamemBERT + SVM} & \textbf{0.923} & 0.959 & \textbf{0.938} & \textbf{0.948} & \textbf{0.866} & 0.908 & \textbf{0.887} \\ 
\midrule
LOGO (Character)    & 0.901 & 0.940 & 0.938 & 0.939 & 0.860 & 0.865 & 0.863 \\ 
LOGO (Author)    & 0.908 & 0.943 & 0.945 & 0.944 & 0.875 & 0.870 & 0.873 \\ 
\midrule
LOGO (10 years)    & 0.915 & 0.950 & 0.943 & 0.946 & 0.872 & 0.886 & 0.879 \\ 
LOGO (20 years)    & 0.904 & 0.940 & 0.943 & 0.942 & 0.870 & 0.865 & 0.867 \\ 
LOGO (50 years)    & 0.892 & 0.928 & 0.952 & 0.940 & 0.885 & 0.832 & 0.858 \\ 
\bottomrule
\end{tabular}
\caption{Detailed models performances: balanced accuracy (B. Acc.), precision (P), recall (R) and F1-score per class for the Detective Detector using different semantic representations and classification algorithms.}
\label{tab:model_results_detailed}
\end{table}


As can be seen in Table~\ref{tab:model_results_detailed}, the LOGO results are minimally different from the best non-LOGO model, leading to the conclusion that despite an unbalanced training corpus in terms of different detectives and authors, the Detective Detector is robust, and not sensitive to the over-representation of some detective figures. 

\begin{figure}[H]
\centering
\includegraphics[width=1\linewidth]{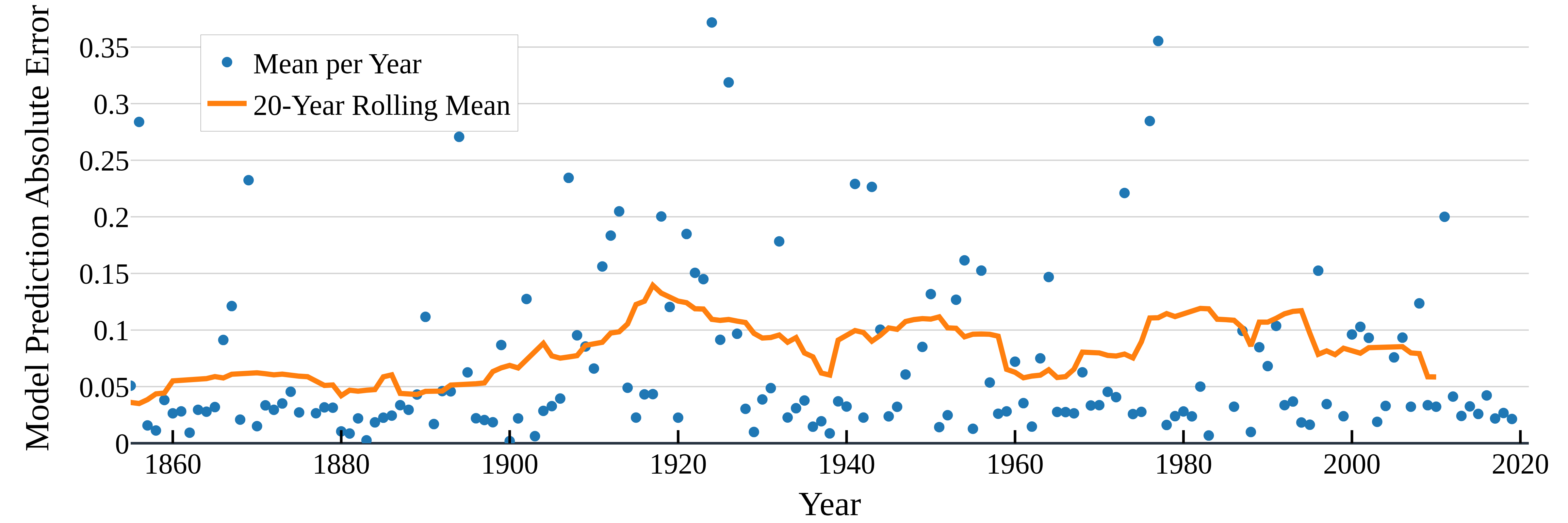}
\caption{Model prediction error over time.}
\label{fig:prediction_error_over_time}
\end{figure}

We carefully examined the temporal stability of the Detective Detector to ensure high scores weren't driven by a specific period and that Detectives were consistently identified across 150 years. To this end, we performed LOGO analyses over 10-, 20-, and 50-year intervals. As shown in Table~\ref{tab:model_results_detailed}, results reveal no significant degradation, indicating the model is not biased toward any particular era. This is further supported by the prediction error over time (Figure~\ref{fig:prediction_error_over_time}), which remains stable despite minor fluctuations. Such diachronic robustness suggests that, beyond stylistic or editorial variation, the investigator figure retains a consistent set of linguistic and narrative features, forming a recognizable discursive imprint that transcends time and unifies 150 years of detective fiction.



\subsection{Qualitative Evaluation of the Distinctive Attributes of the Detective Archetype}\label{qualitative_eval}

While we assessed that the Detective Detector obtains a high accuracy, we also wanted to understand what distinguishes Detective characters from Non-detectives. For this, we look at the attributes (lexical items) that are associated to Detective and Non-detective characters across the whole Chapitres corpus where we detected 713 Detectives and 29,152 Non-Detectives. To compare the attributes of the two types of characters, we used a method based on normalized z-scores in the [-1, 1] interval, where +1 indicates the most distinctive attributes for detectives.

The z-score are calculated in the following manner: the numerator is the log-odds ratio with Dirichlet prior smoothing. The denominator is the standard error estimate based on smoothed counts. This method is based on \textcite{Monroe_Colaresi_Quinn_2017}'s method for informative lexical feature extraction.\\

\noindent
\begin{minipage}[c]{0.4\textwidth}
\LARGE
\[
z_{\text{attr}} = 
\frac{
\log\left(\frac{c_1 + \frac{p}{n}}{n_1 + 1} \middle/ \frac{c_2 + \frac{p}{n}}{n_2 + 1}\right)
}{
\sqrt{\frac{1}{c_1 + \frac{p}{n}} + \frac{1}{c_2 + \frac{p}{n}}}
}
\]
\end{minipage}%
\hfill
\begin{minipage}[c]{0.55\textwidth}
\begin{description}[itemsep=0.1em, parsep=0pt, leftmargin=2.5em]
    \item[$z_{\text{attr}}$:] Attribute distinctiveness score\\(positive = more associated with group 1)
    \item[$c_1$:] Attribute count in group 1 (detectives)
    \item[$c_2$:] Attribute count in group 2 (non--detectives)
    \item[$p$:] Attribute count in the corpus (\( p = c_1 + c_2 \))
    \item[$n_1$:] Sum of all attributes in group 1 (\( n_1 = \sum c_1 \))
    \item[$n_2$:] Sum of all attributes in group 2 (\( n_2 = \sum c_2 \))
    \item[$n$:] Sum of all attributes (\( n = n_1 + n_2 \))
\end{description}
\end{minipage}

\begin{figure}[H]
\centering
\begin{subfigure}[t]{0.4\textwidth}
    \includegraphics[width=\linewidth]{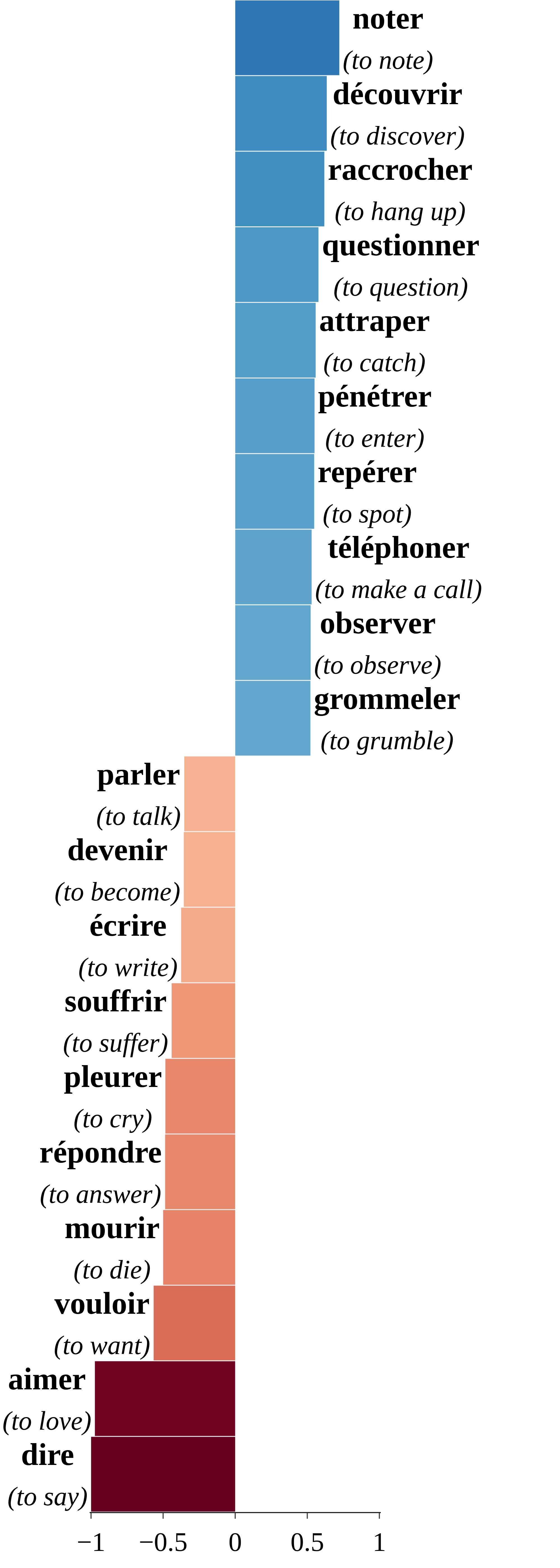}
    \caption{Agent verbs}
\end{subfigure}%
\hspace{-5em} 
\begin{subfigure}[t]{0.4\textwidth}
    \includegraphics[width=\linewidth]{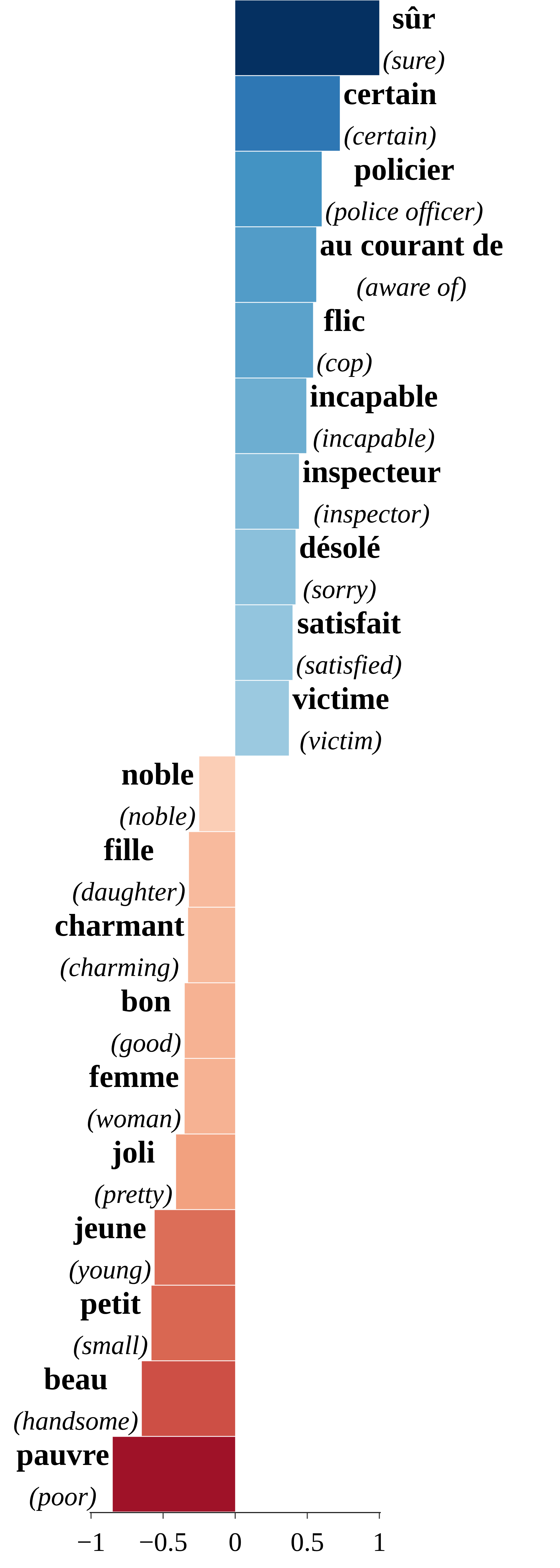}
    \caption{Modifiers}
\end{subfigure}%
\hspace{-5em} 
\begin{subfigure}[t]{0.4\textwidth}
    \includegraphics[width=\linewidth]{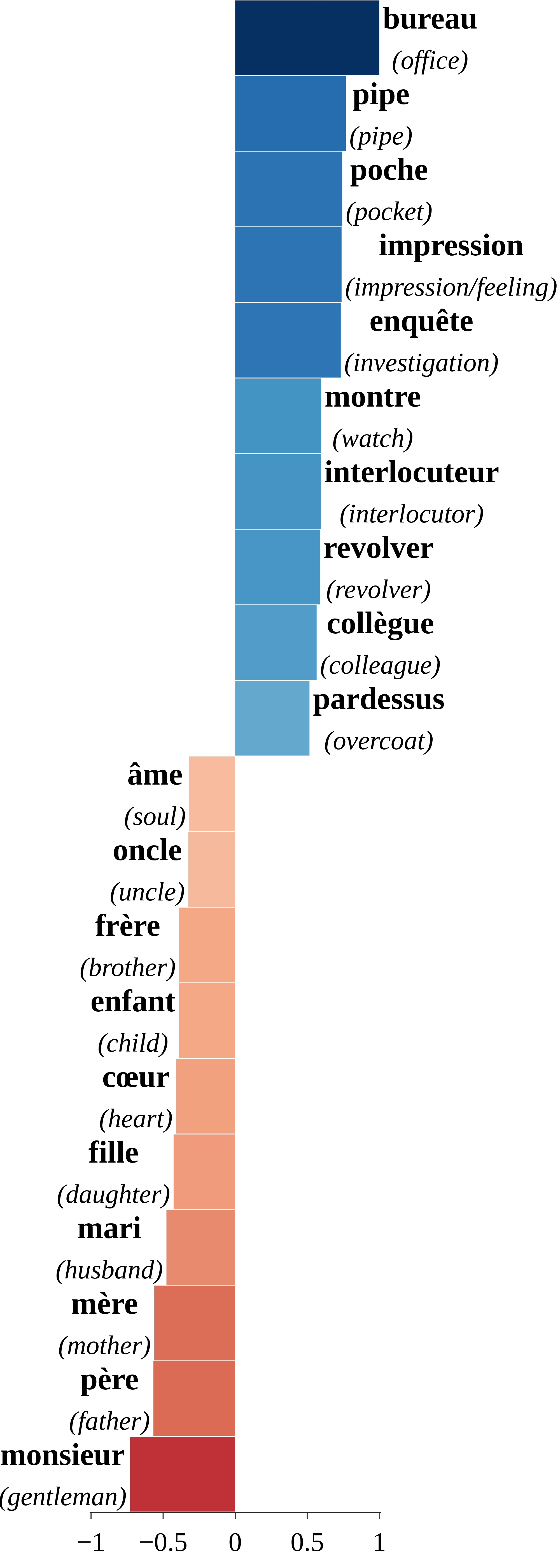}
    \caption{Possessives}
\end{subfigure}
\caption{Attribute distinctiveness of the Detective figure, measured by normalized z-score. A value of \(+1\) indicates the most strongly detective-associated attribute and \(-1\) the least.}
\label{fig:attributes_distinctiveness}
\end{figure}

For the three retained types of attributes (Agentive verbs, Modifiers and Possessives), \autoref{fig:attributes_distinctiveness} shows the vocabulary that is the most typical of the detective archetype (in blue) and the least (in red). The z-scores of agentive verbs get around terms such as \textit{note}, \textit{discover}, \textit{question}, \textit{spot}, and \textit{observe} underscores the detective's fundamental identity as a methodical thinker. These verbs capture a relentless focus on evidence gathering and inference, painting the sleuth as a near-mechanical reasoning apparatus. Equally telling is the stark under-representation of verbs like \textit{love}, \textit{cry}, \textit{suffer}, or \textit{desire}, which reveals an emotional reserve that sets detectives apart from other characters; their narrative presence is defined not by passion or turmoil but by an unwavering, almost clinical detachment.

Turning to possessive constructions, detectives rarely invoke familial bonds or intimate relationships. References to \textit{mother}, \textit{father}, or \textit{daughter} possessions are virtually absent, while their \textit{overcoat}, \textit{pocket}, \textit{revolver} and \textit{watch} abound, which are emblems of a vocation that privileges tools of observation over the ties of kinship. This lexical pattern evokes that iconic image of the lone investigator: coat buttoned high, pockets bulging with instruments of the trade, a revolver at the hip and a timepiece ever-present, signaling both professional rigor and personal solitude.

Finally, the modifiers panel lays bare the cultural coding of the detective as neither female nor ornamental. Highly negative scores for descriptors like \textit{woman}, \textit{girl}, \textit{pretty}, or \textit{handsome} reinforce the image that the detective is rarely described as beautiful or feminine. Instead, the archetype crystallizes as a solitary, intellectually relentless figure whose practical attire and emotionally neutral posture emphasize function over form. Altogether, these lexical signatures confirm and nuance the long-standing critical portrayal of the detective as an isolated, cerebral agent whose world is built on observation, deduction, and the tools of investigation rather than on personal ties or expressive affect.







\section{The Emergence of the Detective Figure}\label{section:emergence_centrality}

To extend our characterization of the Detective Archetype beyond the manually annotated set, we applied the CamemBERT+SVM model ---our best-performing detector---to all characters in the Chapitres corpus. For each novel, we retained the ten most prominent characters (by frequency of mention), yielding a sample of 29,610 characters. 713 of them were classified as Detectives. This large-scale inference allows us to trace both the quantitative rise of detective figure and its narrative prominence over more than a century of French fiction.

\begin{figure}[H]
\centering
\includegraphics[width=1\linewidth]{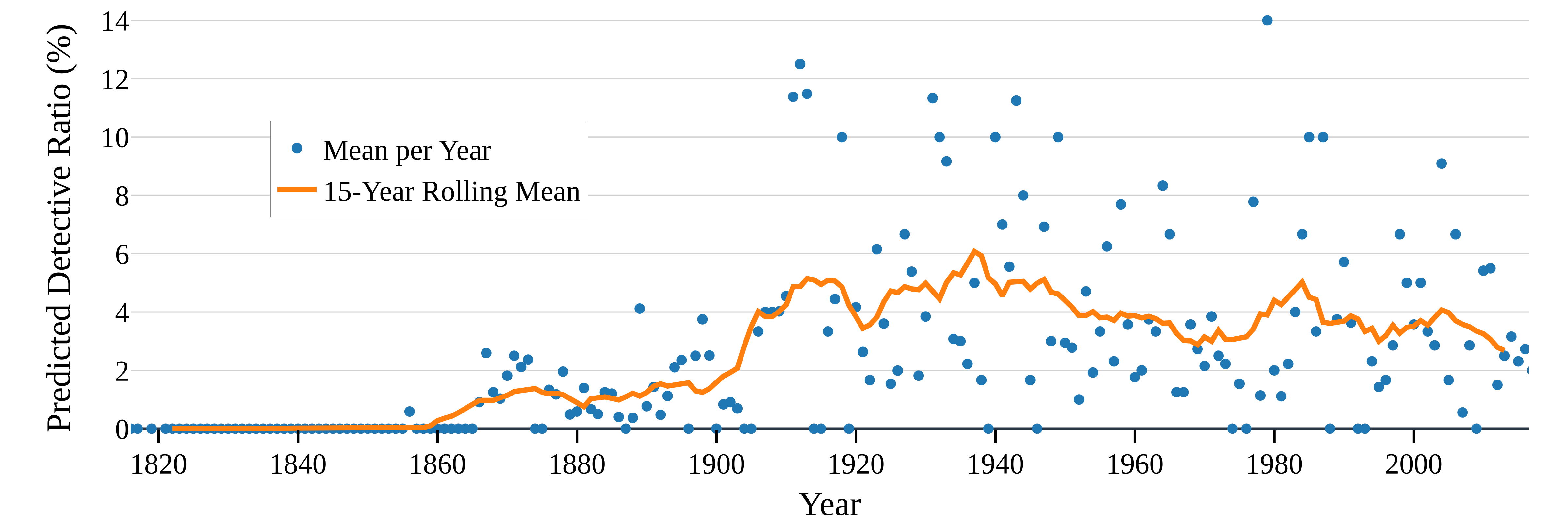}
\caption{Model-predicted detective character ratio over time.}
\label{fig:temporal_predictions}
\end{figure}

\autoref{fig:temporal_predictions} shows that the first literary detective emerge around 1860, with a steep increase in their proportion from 1900 up to nearly 6\% of all characters by the late 1930s. After 1940, the relative frequency of detectives dips to about 4\%, before experiencing a modest revival in the 1980s.

\begin{figure}[H]
\centering
\includegraphics[width=1\linewidth]{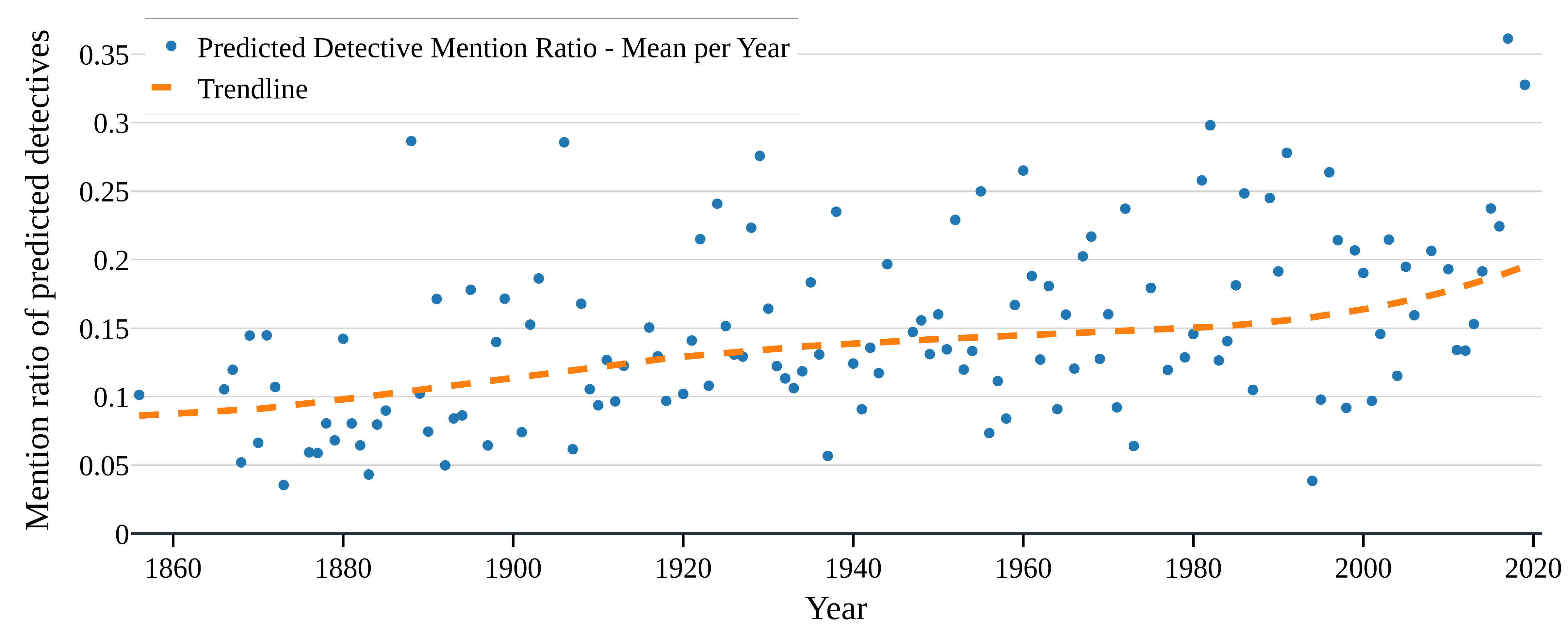}
\caption{Progressive establishment of the detective as the narrative core.}
\label{fig:detective_centrality}
\end{figure}

\textcite{lavergne_naissance_2009} identifies the figure of the investigator as a key element in the emergence of the detective-novel genre. Formerly relegated to a secondary role, the investigator becomes the central character who steers the gradual reconstruction of the drama. We further assessed the detective's narrative centrality via the mention-ratio (the length of a character's coreference chain relative to the book's average). \autoref{fig:detective_centrality} reveals an ongoing upward trajectory from the mid-19th century to the end of our corpus, with a first inflection around 1900. The quadratic fit trendline underscores how, as detective fiction matured, the investigator increasingly anchored the narrative, even as subgenres diversified between 1950 and 2000 (\textit{hard-boiled, série noir, néo-polar}), broadening the range of detective portrayals while preserving their elevated visibility.

The conjunction of these two findings, an increasing number of detective figures and their growing centrality, suggests a dual movement. On the one hand, the archetype crystallizes early enough to remain recognizable even as detective fiction experiments with new forms. On the other hand, the narrative's increasing reliance on its investigator figure indicates that the mystery's resolution depends more and more on the detective's voice and subjectivity; he is no longer a mere auxiliary to the plot, but the pivot around which narrative tension is organized.

This raises a central question for the rest of our study: to what extent does the \textit{voice} or discursive identity of the detective archetype remain constant across 150 years of genre history? Beyond the persistence of a core set of textual markers, is the archetype's semantic imprint (its characteristic ways of speaking, perceiving, and investigating) truly stable, or does it undergo subtle shifts as it traverses successive generic and historical contexts? To address this, we turn to a fine-grained, diachronic analysis of the archetype's semantic trajectory.


\section{The Semantic Trajectory of the Detective Archetype}\label{section:semantic_trajectory}

\begin{figure}[H]
\centering
\includegraphics[width=1\linewidth]{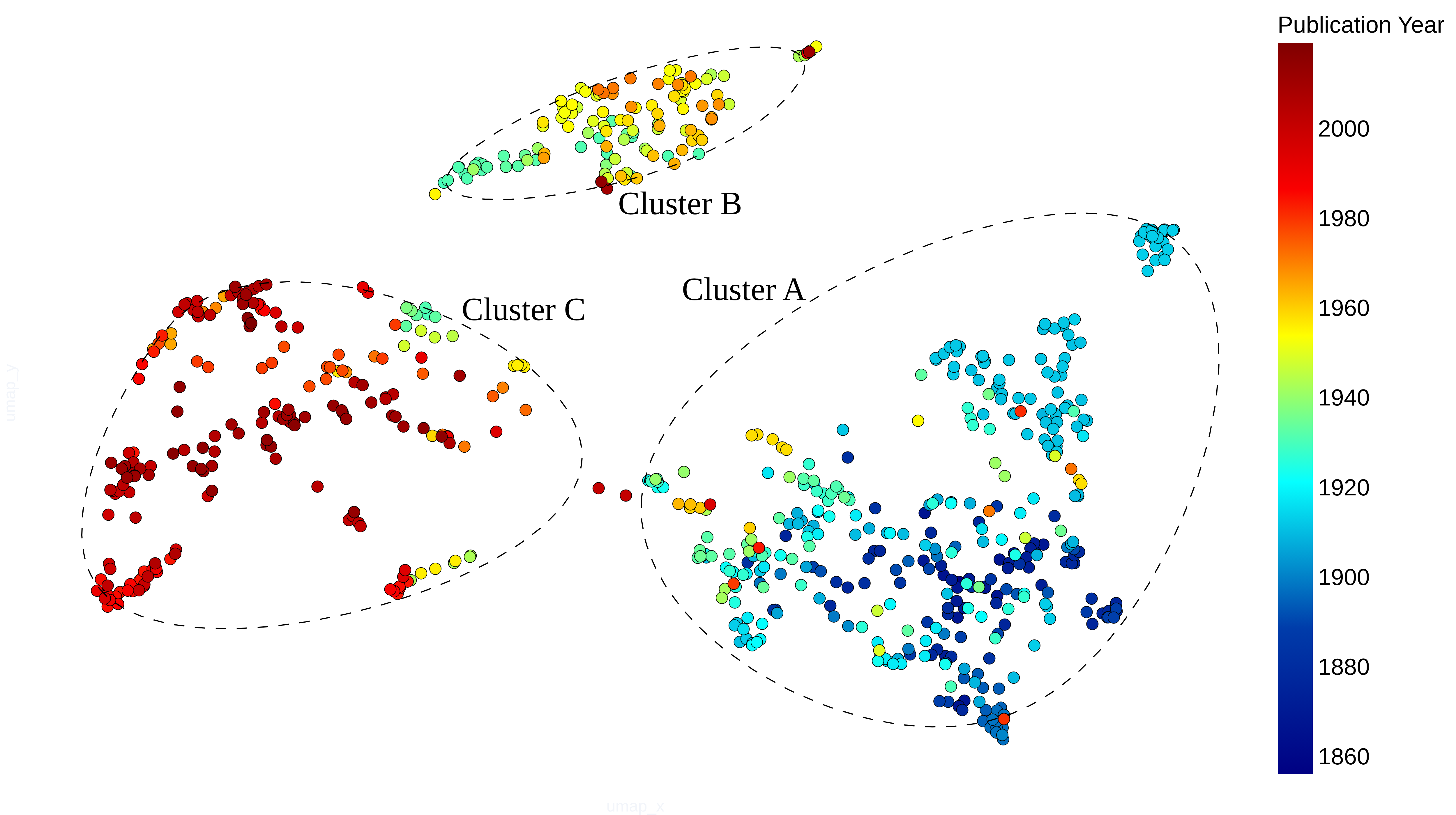}
\caption{Predicted detective clusters.}
\label{fig:detective_clusters}
\end{figure}

In this section, we examine how the detective archetype evolves over time by uncovering subtle semantic shifts beyond its core textual markers. To trace this fine-grained trajectory, we take the CamemBERT contextual embeddings for the 713 characters our model flagged as Detectives in the Chapitres corpus and run a clustering experiment: first reducing these high-dimensional representations to two dimensions using the uniform manifold approximation and projection (UMAP) algorithm \cite{UMAP_2020} (\texttt{n\_neighbors=10}). Visually we could clearly distinguish three big clusters in the space. We therefore choose to continue with a clustering based on the K-means algorithms that identifies three clusters and our intuition about the cluster's boundaries were confirmed. The results are displayed in \autoref{fig:detective_clusters}. We can see clearly the influence of the publication year on the clustering: detectives from similar time periods tends to find themselves in the same clusters. 

To determine whether the three clusters reflect distinct phases in the detective subgenre's semantic trajectory, we applied the z-score method from Subsection~\ref{qualitative_eval} to extract each cluster's most distinctive lexical attributes. This time, we only took in to consideration the positively scores attributes (z-score > 0). We can find this vocabulary in Table~\ref{tab:clusters_most_distinctive_attributes}. We interpret the specific traits of the three clusters by having a closer look at this vocabulary and looking it up in the context of the Chapitre Corpus.


\begin{table}[H]
\centering
\scriptsize  
\begin{tabular}{cccc}
\toprule
 & \textbf{Cluster A} & \textbf{Cluster B} & \textbf{Cluster C} \\
\midrule
\rotatebox{90}{\makecell{\textbf{Agent Verbs}}} & \makecell{dire (to say)\\s'écrier (to exclaim)\\répondre (to answer)\\déclarer (to declare)\\murmurer (to murmur)\\répliquer (to reply)\\interrompre (to interrupt)\\venir (to come)\\demeurer (to remain)\\reprendre (to resume)\\ajouter (to add)\\interroger (to question)\\affirmer (to assert)\\crier (to shout)} & \makecell{questionner (to question)\\téléphoner (to make a call)\\supposer (to suppose)\\regarder (to look)\\bourrer (to stuff)\\préférer (to prefer)\\fumer (to smoke)\\connaître (to know)\\boire (to drink)\\avoir (to have)\\se souvenir (to remember)\\savoir (to know)\\permettre (to allow)\\soupirer (to sigh)} & \makecell{sentir (to feel)\\attraper (to catch)\\poser (to put down)\\foutre (to not care / to make fun of)\\repérer (to spot)\\tenter (to attempt)\\lever (to get up)\\croiser (to run into)\\acquiescer (to nod / to agree)\\se souvenir (to remember)\\sortir (to go out)\\enfiler (to put on (clothes))\\vérifier (to check)\\avaler (to swallow)} \\
\midrule
\rotatebox{90}{\makecell{\textbf{Modifiers}}} & \makecell{cher (dear)\\jeune (young)\\brave (brave)\\malheureux (unhappy)\\vieux (old)\\pauvre (poor)\\excellent (excellent)\\faux (false / fake)\\digne (worthy)\\mystérieux (mysterious)\\général (general)\\bon (good)\\honnête (honest)\\misérable (miserable)} & \makecell{sûr (sure / confident)\\lourd (heavy)\\roux (red-haired)\\peine (sorrow / grief)\\commissaire (commissioner)\\maussade (gloomy / sullen)\\police (police)\\spécial (special)\\ivre (drunk)\\marié (married)\\incapable (incapable)\\fatigué (tired)\\gros (fat)\\large (broad / wide)} & \makecell{désolé (sorry)\\petit (small)\\flic (cop)\\nu (naked)\\responsable (responsible)\\certain (certain)\\seul (alone)\\censé (supposed)\\conscient (aware)\\incapable (incapable)\\type (guy / dude)\\proche (close / nearby)\\enquêteur (investigator)\\au courant de (aware of)} \\
\midrule
\rotatebox{90}{\makecell{\textbf{Possessives}}} & \makecell{ami (friend)\\parole (word / promise)\\compagnon (companion)\\dieu (god)\\maître (master)\\monsieur (gentleman)\\revolver (revolver)\\ordre (order)\\foi (faith)\\cabinet (office / cabinet)\\ennemi (enemy)\\devoir (duty)\\complice (accomplice)\\cheval (horse)} & \makecell{pipe (pipe)\\bureau (office / desk)\\femme (wife)\\pardessus (overcoat)\\chapeau (hat)\\inspecteur (inspector)\\impression (impression / feeling)\\envie (desire)\\interlocuteur (interlocutor)\\collègue (colleague)\\veston (jacket)\\collaborateur (collaborator)\\poche (pocket)\\client (client)} & \makecell{veste (jacket)\\corps (body)\\téléphone (phone)\\sac (bag)\\mal (pain / suffering)\\lunette (glasses)\\voiture (car)\\dos (back)\\père (father)\\visage (face)\\ventre (belly / stomach)\\doigt (finger)\\peau (skin)\\mère (mother)}\\
\bottomrule
\end{tabular}
\caption{Detective clusters most distinctive attributes.}
\label{tab:clusters_most_distinctive_attributes}
\end{table}


Cluster A corresponds to the “classical” rational detective of the late nineteenth and early twentieth centuries: figures who resolve mysteries primarily through methodical deduction and interrogation, as in the serialized fiction of Gaboriau, Leroux, and Leblanc~\parencite{vareille_roman_1985}. In addition to high z-scores for agentive verbs like \textit{question}, this cluster is distinguished by an abundance of dialogue markers: \textit{say}, \textit{answer}, \textit{declare}, \textit{reply}, reflecting the detective's reliance on spoken questioning and verbal exchanges. Unlike the more introspective or action-oriented patterns in the other clusters, Cluster A's signature lies in its emphasis on the detective as an interlocutor whose investigative power is enacted through dialogue.


Cluster B, by contrast, moves beyond this cool rationality to embrace a more humanized investigator figure. It aligns with France's Golden Age of detective fiction, symbolically initiated in 1927 by the \textit{Masque} collection\footnote{See \textcite{martinetti_masque_1997} for the history of the collection.} and notably symbolized by the emergence of Simenon's Inspector Maigret in the 1930s, whose introspective empathy and attention to social detail depart from the archetype's earlier detachment. Here, lexical markers shift toward verbs and modifiers that convey compassion, observation of daily life, and a nuanced portrayal of the detective's own emotional landscape.

We can observe this by the fact that words as \textit{soupirer} (to sigh) and \textit{supposer} (to suppose) are makers of Cluster B (see Examples~\ref{ex:maigret1} and~\ref{ex:maigret2}). 

\ex. \label{ex:maigret1}\a. Maigret, \textbf{soupira}, prit un temps pour allumer sa pipe.\footnote{Commissioner Jules Maigret from \textit{Le Port des brumes} by Georges Simenon (1932).}
\b. Maigret \textbf{sighed} and took a moment to light his pipe.

\ex.\label{ex:maigret2} \a. Maigret \textbf{supposa} que Joseph avait amené Jaja et Sylvie à Antibes. Il ne se trompait pas.\footnote{Commissioner Jules Maigret from \textit{Liberty Bar} by Georges Simenon (1932).}
\b. Maigret \textbf{guessed} that Joseph had brought Jaja and Sylvie to Antibes. He was not mistaken.

As noted by Cawelti, \enquote{Maigret's skills as a detective [...] are not ratiocinative but a result of tireless energy and irrational intuition} \cite[126]{cawelti_adventure_1997}. Simenon's work transforms the detective narrative into \textit{a police procedural}, moving away from the traditional \textit{whodunit}: \enquote{there is no mystery in the inverted tale about who killed the victim and how; the mystery is whether the detective will be able to solve the crime} \cite[126]{cawelti_adventure_1997}. No longer a distant genius, Maigret becomes a profoundly human figure: \enquote{the flow of the narrative is largely confined to the flow of Maigret's perceptions, observations, and feelings} \cite[127]{cawelti_adventure_1997}, and the reader is therefore \enquote{privy to many of the inferences and deductions he draws from the clues} \cite[127]{cawelti_adventure_1997}.

Cluster C represents the modern period, emerging post-1945 with the introduction of the American \textit{hardboiled} style in France. However, this period also sees the continued evolution of the empathetic detective tradition. Fred Vargas's Commissaire Adamsberg (first appearing in 1991) exemplifies an alternative modern trajectory: \enquote{Unlike most data-driven detectives who center their conclusions on deliberation of facts and figures, Adamsberg, the sensitive semiotician, clearly resides in the post-World War I era where \enquote{new methods of investigation [based] on instinct, intuition, and empathy replaced rational deduction}} \cite{becker_georges_1977}. 

Yet alongside such introspective figures, the \enquote{American moment of French literature}, as described by \textcite{cadin_moment_2018}, fundamentally reshaped French detective fiction with the emergence of the \textit{Série Noire} and, later, the \textit{néo-polar}. Authors such as Boris Vian, Léo Malet (with \textit{Nestor Burma}), and Didier Daeninckx (with \textit{Gabriel Lecouvreur}) adopted this style, creating investigations that were far more realistic, physically dangerous, and marked by a certain cynicism. The detective is no longer simply a rational and observant thinker, but became a character immersed in social violence, often disillusioned, and navigating a dark world where the line between good and evil was increasingly blurred.

In the vocabulary, we identified for example the word \textit{foutre}, which is a vulgar polysemous word that originally means \textit{to fuck}, but is used in modern language as a swear word for different matters, for example in the saying \textit{Je m'en fous.}, meaning something such as \textit{I don't give a shit.} and \textit{Qu'est-ce que vous foutez ?} meaning \textit{What the hell are you doing?}. The vulgarity of the language is typical for the violence in the subgenre (see Example~\ref{ex:foutre} for an illustration). 

\ex. \label{ex:foutre} \a. Le visage du commandant Verhoeven change brusquement, on ne rigole plus. Il plaque violemment le téléphone sur la table en fer. - Et maintenant, tu me \textbf{fous} un bordel noir dans la communauté. Je veux une fille, vingt-cinq-trente ans, pas mal mais crevée. Sale.\footnote{Commander Verhoeven from \textit{Alex} by Pierre Lemaitre (2011).}
\b. The face of Commander Verhoeven changes abruptly - no more joking around. He slams the phone violently onto the metal table. - "And now you're \textbf{fucking everything up} in the community. I want a girl, twenty-five to thirty years old, decent-looking but worn out. Dirty."

From the vocabulary we identify a detective that is more part of the action, for example the verbs \textit{enfiler} (to put on) and \textit{croiser}
(to run into) are distincitve of Cluster C. 

\ex. \a. Camille comprend, se lève, \textbf{enfile} son manteau, prend son chapeau et sort. Au passage, il \textbf{croise} Armand.\footnote{Camille from \textit{Alex} by Pierre Lemaitre (2011).}
\b. Camille understands, gets up, \textbf{puts on} his coat, takes his hat, and goes out. On the way, he \textbf{runs into} Armand.



%




\section{Discussion}






\subsection{Literary Interpretation of Results}
Our diachronic prediction across nearly 150 years of francophone detective fiction demonstrates the remarkable strength and durability of the detective archetype as the linchpin of the narrative. From its earliest affirmation, the detective has remained the central figure around whom the story pivots, resisting the vicissitudes of time even as its expressive form evolves. Although the core investigatory function (unraveling mysteries through observation and inference) remains intact, we observe a progressive enrichment of the archetype. What began as a cold, efficient “reasoning machine” gradually acquires a fuller emotional envelope and a deeper engagement with the world “on the ground,” reflecting changing narrative priorities and reader expectations.

The bottom-up clustering analysis yields three distinct poles that correspond to major historical phases, each delineated by a significant temporal shift. The clarity of this tripartition might not have been obvious from traditional literary scholarship alone, yet it emerges naturally from the data: an early rational-puzzle phase, a mid-century emotionally engaged phase, and a contemporary hard-edged phase characterized by moral ambiguity. These three clusters make sense not only as statistical artifacts but also as literary realities, each shaped by broader socio-cultural and aesthetic transformations in French crime writing. These three poles are not merely artifacts of chronological periodization but resonate deeply with literary function and reader positioning. In the rational-puzzle phase, the reader is an intellectual spectator; in the empathic-procedural phase, an accomplice to the detective's intuition; and in the hardboiled-moral-ambiguous phase, a witness to the detective's struggle to impose order on a chaotic society. This development reflects broader socio-literary transformations in French detective fiction, transitioning from narratives of pure detection to novels of criminality, victimhood, and ultimately violence \parencite{tourteau_arsene_1970}. As the genre evolved to mirror shifting societal attitudes, values, and political realities, the detective archetype itself necessarily adapted, navigating toward representations of social violence and moral ambiguity.




\subsection{Methodological Limitations and Interpretative Challenges}

One limitation of our study is its reliance on the off-the-shelf BookNLP-fr pipeline, via its Propp extension, for character identification, coreference resolution, and attribute extraction. Though designed for long-form French literature, the pipeline introduces uncertainty, as our analysis is bound by its performance and error patterns. This affects coreference accuracy in particular: the system may mistakenly merge distinct characters or split one into several chains. Such errors introduce noise, potentially distorting character representations. While we thoroughly assessed our classification models and attribute relevance, the dataset's scale prevents full manual validation of coreference chains. Consequently, some detective profiles may include attributes from unrelated characters, impacting both archetype identification and its diachronic analysis.

Another limitation lies in our attribute extraction method, which relies on lemmatized unigrams. While this reduces vocabulary size and facilitates analysis, it also introduces key drawbacks. The approach does not account for negation (for instance, treating “intelligent” and “not intelligent” as equivalent) which can invert or alter meaning. It also overlooks gradation (e.g., “very clever” vs. “somewhat clever”) and the broader context in which attributes appear. Furthermore, polysemous terms are treated in isolation, without disambiguation, which risks conflating distinct meanings and misrepresenting characters. Addressing these issues would require more advanced linguistic processing, such as incorporating multi-word expressions, explicitly handling negation, and using context-aware embeddings or disambiguation models to improve the precision and nuance of character profiling.

\section{Conclusion}
This paper has proposed a novel method for tracking character archetypes within literary genres over time. We have applied this approach to the figure of the detective, a particularly well-suited case for such an investigation. 
Our results show a remarkable degree of stability in the textual signature associated with the detective archetype. An analysis of the linguistic features most strongly associated with the detective role confirms the persistence of a core investigative lexicon, comprising characteristic modifiers and verbs across the corpus. Yet, a more detailed examination of model outputs reveal both the figure's permeability and historical contingency. Newer subgenres introduce additional elements, including subjectivity, cynicism, and existential doubt. This evolution is consistent with theoretical perspectives that frame the detective not as a fixed type, but as a cultural site for negotiating the tensions between reason, intuition, and moral ambiguity in changing historical contexts.

Future research may benefit from moving beyond binary classifications (Detective vs. Non-detective) to explore a spectrum of detectiveness, as well as its interaction with other character roles. Such an approach could further elucidate the detective's dual status as both a narrative anchor and a locus of transformation within the genre. 
In addition, this methodology opens new avenues for examining archetypal characters across other genres. It enables a dual perspective: not only can we track the defining traits of central characters, but we can also investigate points of similarity and overlap among secondary figures. By integrating \textit{distant reading} techniques (such as large-scale feature extraction and clustering) with \textit{close reading} of selected examples, we aim to identify salient features, assess their interpretability, and situate them within broader literary traditions. We anticipate that further applications of this method across diverse genres will offer valuable insights into character typology and contribute to the refinement (or reassessment) of existing literary theories.

\section*{Acknowledgements}
This research was funded in part by PRAIRIE-PSAI (Paris Artificial intelligence Research institute–Paris School of Artificial Intelligence, reference ANR-22-CMAS-0007). This work has received support under the Major Research Program of PSL Research University "CultureLab" launched by PSL Research University and implemented by ANR with the references ANR-10-IDEX-0001.

\printbibliography

@book{becker_georges_1977,
	address = {Boston, Mass},
	series = {Twayne's world authors series {France}},
	title = {Georges {Simenon}},
	isbn = {978-0-8057-6293-8},
	language = {eng},
	number = {456},
	publisher = {Twayne},
	author = {Becker, Lucille F.},
	year = {1977},
}

@inproceedings{barre_latent_2024,
	location = {Aarhus},
	title = {Latent Structures of Intertextuality in French Fiction},
	rights = {Creative Commons Attribution 4.0 International},
	eventtitle = {{CHR} 2024},
	pages = {21--26},
	booktitle = {Proceedings of the Computational Humanities Research Conference 2024},
	author = {Barré, Jean},
	date = {2024},
    url={https://ceur-ws.org/Vol-3834/paper97.pdf},
}

@article{scikit-learn,
  title={Scikit-learn: Machine Learning in {P}ython},
  author={Pedregosa, F. and Varoquaux, G. and Gramfort, A. and Michel, V.
          and Thirion, B. and Grisel, O. and Blondel, M. and Prettenhofer, P.
          and Weiss, R. and Dubourg, V. and Vanderplas, J. and Passos, A. and
          Cournapeau, D. and Brucher, M. and Perrot, M. and Duchesnay, E.},
  journal={Journal of Machine Learning Research},
  volume={12},
  pages={2825--2830},
  year={2011}
}

@article{Pandzic_2020, title={E. A. Poe and A. A. Shkljarevskij: Foregrounding Deduction and/or Social Commentary – A Comparative Study of Early Detective Fiction}, DOI={10.15291/sic/1.11.lc.5}, number={1.11}, journal={[sic] - a journal of literature, culture and literary translation}, publisher={University of Zadar}, author={Pandžić, Maja}, year={2020}, month=dec}

@incollection{schutt_rivalry_1998,
  author    = {Schutt, Sita},
  title     = {Rivalry and Influence: Nineteenth-Century French Detective Narratives},
  booktitle = {The Art of Murder},
  editor    = {Gustav Klaus and Stephen Knight},
  year      = {1998},
  pages     = {38--49}
}

@book{lycett_man_2007,
	address = {New York},
	edition = {1st Free Press hardcover ed},
	title = {The man who created {Sherlock} {Holmes}: the life and times of {Sir} {Arthur} {Conan} {Doyle}},
	isbn = {978-0-7432-7523-1},
	shorttitle = {The man who created {Sherlock} {Holmes}},
	publisher = {Free Press},
	author = {Lycett, Andrew},
	year = {2007},
}

@book{bonniot_emile_1985,
	address = {Paris},
	title = {Émile {Gaboriau} ou {La} naissance du roman policier},
	isbn = {978-2-7116-9277-4},
	language = {fre},
	publisher = {Vrin},
	author = {Bonniot, Roger},
	year = {1985},
}

@inproceedings{ehrmanntraut-etal-2023-llpro,
    title = "{LL}pro: A Literary Language Processing Pipeline for {G}erman Narrative Texts",
    author = "Ehrmanntraut, Anton  and
      Konle, Leonard  and
      Jannidis, Fotis",
    editor = "Georges, Munir  and
      Herygers, Aaricia  and
      Friedrich, Annemarie  and
      Roth, Benjamin",
    booktitle = "Proceedings of the 19th Conference on Natural Language Processing (KONVENS 2023)",
    month = sep,
    year = "2023",
    address = "Ingolstadt, Germany",
    publisher = "Association for Computational Lingustics",
    url = "https://aclanthology.org/2023.konvens-main.3/",
    pages = "28--39"
}

@inbook{ALL_detective_2012,
  author    = {G.J. van Bork and D. Delabastita and H. Van Gorp and P.J. Verkruijsse and G.J. Vis and Lars Bernaerts and Frank Willaert and Esther Op de Beek and Nina Geerdink and Sara Van den Bossche and Orsolya Réthelyi},
  title     = {Algemeen Letterkundig Lexicon},
  chapter   = {Detectiveroman},
  pages     = {n.p.},
  year      = {2012},
  publisher = {Digitale Bibliotheek voor de Nederlandse Letteren (DBNL)},
  address   = {Den Haag},
  url       = {https://www.dbnl.org/tekst/dela012alge01_01/dela012alge01_01_00775.php},
}

@inproceedings{naguib2022romanciers,
  title={Romanciers et romanci{\`e}res du XIX{\`e}me si{\`e}cle: une {\'e}tude automatique du genre sur le corpus GIRLS (Male and female novelists: an automatic study of gender of authors and their characters)},
  author={Naguib, Marco and Delaborde, Marine and Andrault, Blandine and Bekolo, Ana{\"\i}s and Seminck, Olga},
  booktitle={Actes de la 29e Conf{\'e}rence sur le Traitement Automatique des Langues Naturelles. Atelier TAL et Humanit{\'e}s Num{\'e}riques (TAL-HN)},
  pages={66--77},
  year={2022}
}

@article{moretti2000conjectures,
  title={Conjectures on world literature},
  author={Moretti, Franco},
  journal={New left review},
  volume={2},
  number={1},
  pages={54--68},
  year={2000}
}

@book{moretti2013,
	title = {Distant Reading},
    publisher = {Verso},
	author = {Moretti, Franco},
    year = {2013},
	address = {London},
}

@mastersthesis{murray_thesis_on_Poe,
  title        = {Edgar Allan Poe and Science: Unraveling the Plot of the Universe},
  author       = {Ellison, Murray S},
  year         = 2015,
  note         = {Available at \url{https://scholarscompass.vcu.edu/cgi/viewcontent.cgi?article=5047&context=etd}},
  school       = {Virginia Commonwealth University},
  type         = {Master's thesis}
}

@incollection{williard_huntington_wright_great_1947,
	address = {New York},
	title = {The {Great} {Detective} {Stories}},
	booktitle = {The {Art} of the {Mystery} {Story}},
	publisher = {Grosset \& Dunlap},
	author = {{Williard Huntington Wright}},
	year = {1947},
	pages = {33--70},
}

@inproceedings{bamman_bayesian_2014,
	address = {Baltimore, Maryland},
	title = {A {Bayesian} {Mixed} {Effects} {Model} of {Literary} {Character}},
	doi = {10.3115/v1/P14-1035},
	urldate = {2023-04-10},
	booktitle = {Proceedings of the 52nd {Annual} {Meeting} of the {Association} for {Computational} {Linguistics} ({Volume} 1: {Long} {Papers})},
	publisher = {Association for Computational Linguistics},
	author = {Bamman, David and Underwood, Ted and Smith, Noah A.},
	month = jun,
	year = {2014},
	pages = {370--379},
}

@article{underwood_transformation_2018,
	title = {The {Transformation} of {Gender} in {English}-{Language} {Fiction}},
	volume = {3},
	doi = {10.22148/16.019},
	language = {en},
	number = {2},
	urldate = {2022-10-06},
	journal = {Journal of Cultural Analytics},
	author = {Underwood, Ted and Bamman, David and Lee, Sabrina},
	month = feb,
	year = {2018},
}

@inproceedings{vianne_gender_2023,
  title        = {Gender {Bias} in {French} {Literature}},
  author = {Vianne, Laurine and Dupont, Yoann and Barré, Jean},
  year         = 2023,
  booktitle    = {Proceedings of the Computational Humanities Research Conference},
  series       = {CEUR Workshop Proceedings},
  volume       = 3558,
  pages        = {247-262},
  editor       = {rtjoms Šeļa and Fotis Jannidis and Iza Romanowska}
}

@inbook{konle_modeling_2022,
  title = {Modeling Plots of Narrative Texts as Temporal Graphs},
  author = {Leonard Konle and Fotis Jannidis},
  year = {2022},
  month = dec,
  language = {English},
  volume = {3304},
  pages = {318--330},
  booktitle = {Proceedings of the Computational Humanities Research Conference (CHR 2022)},
  publisher = {CEUR Workshop Proceedings},
}

@inbook{konle_character_2023,
  title = {On Character Perception and Plot Structure of German Romance Novels},
  author = {Leonard Konle and Agnes Hilger and Fotis Jannidis},
  year = {2023},
  month = dec,
  language = {English},
  volume = {3518},
  pages = {592--605},
  booktitle = {Proceedings of the Computational Humanities Research Conference (CHR 2023)},
  publisher = {CEUR Workshop Proceedings},
}

@phdthesis{cadin_moment_2018,
	address = {Paris},
	title = {Le moment américain du roman français: 1945-1950},
	shorttitle = {Le moment américain du roman français},
	language = {fre},
	school = {Classiques Garnier},
	author = {Cadin, Anne},
	year = {2018},
	isbn = {9782406077510},
	series = {Études de littérature des XXe et XXIe siècles},
	number = {71},
}

@article{martin2019camembert,
  title={CamemBERT: a tasty French language model},
  author={Martin, Louis and Muller, Benjamin and Su{\'a}rez, Pedro Javier Ortiz and Dupont, Yoann and Romary, Laurent and de La Clergerie, {\'E}ric Villemonte and Seddah, Djam{\'e} and Sagot, Beno{\^\i}t},
  journal={arXiv preprint arXiv:1911.03894},
  year={2019}
}

@article{dubois_naissance_1985,
	title = {Naissance du récit policier},
	volume = {60},
	copyright = {free},
	doi = {10.3406/arss.1985.2287},
	language = {fre},
	number = {1},
	urldate = {2025-05-16},
	journal = {Actes de la Recherche en Sciences Sociales},
	author = {Dubois, Jacques},
	year = {1985},
	pages = {47--55},
}

@book{cawelti_adventure_1997,
	address = {Chicago, Ill.},
	edition = {pbk. ed., [Nachdr.]},
	title = {Adventure, mystery, and romance: formula stories as art and popular culture},
	isbn = {978-0-226-09867-8},
	shorttitle = {Adventure, mystery, and romance},
	language = {eng},
	publisher = {The Univ. of Chicago Press},
	author = {Cawelti, John G.},
	year = {1997},
}

@book{brownson_figure_2014,
	address = {Jefferson},
	title = {The figure of the detective: a literary history and analysis},
	isbn = {978-1-4766-1272-0},
	shorttitle = {The figure of the detective},
	language = {eng},
	publisher = {McFarland},
	author = {Brownson, Charles},
	year = {2014},
}

@incollection{vareille_roman_1985,
	address = {Lyon},
	series = {Littérature \&amp; idéologies},
	title = {Roman policier archaïque et aventure archaïque},
	copyright = {https://www.openedition.org/12554},
	isbn = {978-2-7297-0999-0},
	language = {fr},
	urldate = {2025-05-14},
	booktitle = {L'{Aventure} dans la littérature populaire au xixe siècle},
	publisher = {Presses universitaires de Lyon},
	author = {Vareille, Jean-Claude},
	editor = {Bellet, Roger},
	year = {1985},
	doi = {10.4000/books.pul.1232},
	note = {Code: L'Aventure dans la littérature populaire au xixe siècle},
	pages = {185--196},
}

@book{underwood2019distant,
  title={Distant horizons: digital evidence and literary change},
  author={Underwood, Ted},
  year={2019},
  publisher={University of Chicago Press}
}

@inproceedings{van2023putting,
  title={Putting Dutchcoref to the Test: Character Detection and Gender Dynamics in Contemporary Dutch Novels},
  author={van Zundert, Joris and van Cranenburgh, Andreas and Smeets, Roel},
  booktitle={Computational Humanities Research Conference},
  pages={757--771},
  year={2023},
  organization={CEUR Workshop Proceedings (CEUR-WS. org)}
}

@inproceedings{bamman-etal-2020-annotated,
    title = "An Annotated Dataset of Coreference in {E}nglish Literature",
    author = "Bamman, David  and
      Lewke, Olivia  and
      Mansoor, Anya",
    editor = "Calzolari, Nicoletta  and
      B{\'e}chet, Fr{\'e}d{\'e}ric  and
      Blache, Philippe  and
      Choukri, Khalid  and
      Cieri, Christopher  and
      Declerck, Thierry  and
      Goggi, Sara  and
      Isahara, Hitoshi  and
      Maegaard, Bente  and
      Mariani, Joseph  and
      Mazo, H{\'e}l{\`e}ne  and
      Moreno, Asuncion  and
      Odijk, Jan  and
      Piperidis, Stelios",
    booktitle = "Proceedings of the Twelfth Language Resources and Evaluation Conference",
    month = may,
    year = "2020",
    address = "Marseille, France",
    publisher = "European Language Resources Association",
    pages = "44--54",
    language = "English",
    ISBN = "979-10-95546-34-4",
}

@misc{github_bamman,
  author = {Bamman, David},
  title = {{BookNLP}},
  url = {https://github.com/booknlp/booknlp},
  year = {2021}
}

@article{booknlp_fr_2024,
	author = {Frédérique Mélanie-Becquet AND Jean Barré AND Olga Seminck AND Clément Plancq AND Marco Naguib AND Martial Pastor AND Thierry Poibeau},
	title = {BookNLP-fr, the French Versant of BookNLP. A Tailored Pipeline for 19th and 20th Century French Literature},
	volume = {3},
	year = {2024},
	issue = {1},
	doi = {10.48694/jcls.3924},
	month = {11},
	pages = {1-34},
	issn = {2940-1348},
	publisher ={Universitäts- und Landesbibliothek Darmstadt},
	journal = {Journal of Computational Literary Studies}
}

@misc{leblond_2022_7446728,
  author       = {Leblond, Aude},
  title        = {Corpus Chapitres},
  year         = {2022},
  publisher    = {Zenodo},
  version      = {v1.0.0},
  doi          = {10.5281/zenodo.7446728},
}

@book{martinetti_masque_1997,
	address = {Amiens},
	series = {Références},
	title = {Le {Masque}: histoire d'une collection},
	isbn = {978-2-906389-82-3},
	shorttitle = {Le {Masque}},
	number = {3},
	publisher = {Encrage},
	author = {Martinetti, Anne},
	year = {1997},
}

@book{lavergne_naissance_2009,
	address = {Paris},
	series = {Études de littérature des {XXe} et {XXIe} siècles},
	title = {La naissance du roman policier français: du {Second} {Empire} à la {Première} {Guerre} mondiale},
	isbn = {978-2-8124-0028-5},
	shorttitle = {La naissance du roman policier français},
	number = {7},
	publisher = {Classiques Garnier},
	author = {Lavergne, Elsa de},
	year = {2009},
	note = {OCLC: ocn436637927},
}

@book{messac_detective_2011,
	address = {Amiens},
	series = {Travaux},
	title = {Le detective novel et l'influence de la pensée scientifique},
	isbn = {978-2-251-74246-5},
	language = {fre},
	number = {55},
	publisher = {Encrage},
	author = {Messac, Régis},
	year = {2011},
	origyear = {1929},         % original publication
}

@incollection{todorov_typologie_1980,
	address = {Paris},
	series = {Points {Littérature}},
	title = {Typologie du roman policier},
	isbn = {978-2-02-005693-9},
	language = {fre},
	number = {120},
	booktitle = {Poétique de la prose},
	publisher = {Éd. du Seuil},
	author = {Todorov, Tzvetan},
	year = {1980},
}

@book{symons_bloody_1993,
	address = {New York},
	edition = {3., rev. ed},
	title = {Bloody murder: from the detective story to the crime novel},
	isbn = {978-0-89296-496-3},
	shorttitle = {Bloody murder},
	language = {eng},
	publisher = {Mysterious Pr},
	author = {Symons, Julian},
	year = {1993},
}

@book{tourteau_arsene_1970,
	title = {D'{Arsène} {Lupin} à {San}-{Antonio}: {Le} roman policier français de 1900 à 1970},
	shorttitle = {D'{Arsène} {Lupin} à {San}-{Antonio}},
	language = {Français},
	publisher = {FeniXX réédition numérique},
	author = {Tourteau, Jean-Jacques},
	month = jan,
	year = {1970},
}

@book{gerson_vidocq_1977,
	address = {Boston},
	title = {The {Vidocq} dossier: the story of the world's first detective},
	isbn = {978-0-395-25176-8},
	shorttitle = {The {Vidocq} dossier},
	language = {eng},
	publisher = {Houghton Mifflin},
	author = {Gerson, Noel B.},
	year = {1977},
}

@article{barre_operationalizing_2023,
	author = {Barr{\' e}, Jean and Camps, Jean-Baptiste and Poibeau, Thierry},
	journal = {Journal of Cultural Analytics},
	doi = {10.22148/001c.88113},
	number = {3},
	year = {2023},
	title = {Operationalizing {Canonicity}: A {Quantitative} {Study} of {French} 19th and 20th {Century} {Literature}},
	volume = {8},
}

@inproceedings{Barré2023,
  TITLE = {{Pour une d{\'e}tection automatique de l'espace textuel des personnages romanesques}},
  AUTHOR = {Barr{\'e}, Jean and Cabrera Ram{\'i}rez, Pedro and M{\'e}lanie, Fr{\'e}d{\'e}rique and Galleron, Ioanna},
  URL = {https://hal.science/hal-04105537},
  BOOKTITLE = {{Humanistica 2023}},
  ADDRESS = {Gen{\`e}ve, Switzerland},
  ORGANIZATION = {{Association francophone des humanit{\'e}s num{\'e}riques}},
  SERIES = {Corpus},
  YEAR = {2023},
  MONTH = Jun,
  HAL_ID = {hal-04105537},
  HAL_VERSION = {v2},
}

@article{Monroe_Colaresi_Quinn_2017,
title={Fightin’ Words: Lexical Feature Selection and Evaluation for Identifying the Content of Political Conflict},
volume={16}, 
DOI={10.1093/pan/mpn018},
number={4},
journal={Political Analysis},
author={Monroe, Burt L. and Colaresi, Michael P. and Quinn, Kevin M.}, year={2017}, 
pages={372–403}}

@misc{UMAP_2020,
      title={UMAP: Uniform Manifold Approximation and Projection for Dimension Reduction}, 
      author={Leland McInnes and John Healy and James Melville},
      year={2020},
      eprint={1802.03426},
      archivePrefix={arXiv},
      primaryClass={stat.ML},
}

@misc{Bourgois2025,
      title={The Elephant in the Coreference Room: Resolving Coreference in Full-Length French Fiction Works}, 
      author={Antoine Bourgois and Thierry Poibeau},
      year={2025},
      eprint={2510.15594},
      archivePrefix={arXiv},
      primaryClass={cs.CL},
    %   url={https://arxiv.org/abs/2510.15594}, 
}

\end{document}